\documentclass[letterpaper, 10 pt, journal, twoside]{IEEEtran}

\usepackage{cite}
\usepackage{graphicx} 
\usepackage{bbm}
\usepackage{algorithm}
\usepackage{algorithmicx}
\usepackage{algpseudocode}
\usepackage{amsmath}
\usepackage{mathrsfs}
\usepackage{multirow}
\usepackage{booktabs}
\usepackage{makecell}
\usepackage{array}
\usepackage{subfigure} 
\usepackage{graphicx}
\usepackage{amssymb}
\usepackage{mathrsfs}
\usepackage{amsfonts,bm}
\usepackage{algorithm}
\usepackage{color}
\usepackage{cite}
\usepackage{graphics} 
\usepackage{epsfig} 
\usepackage{mathptmx} 
\usepackage{times} 
\usepackage{algorithm}
\usepackage{algorithmicx}
\usepackage{setspace}
\usepackage{algpseudocode}
\usepackage{graphicx}
\usepackage{multirow}
\usepackage{rotating}
\usepackage{makecell}
\usepackage{adjustbox}
\usepackage{rotating}
\usepackage[utf8]{inputenc} 
\usepackage[T1]{fontenc}    
\usepackage{hyperref}       
\usepackage{url}            
\usepackage{booktabs}      
\usepackage{nicefrac}       
\usepackage{microtype}      
\usepackage{lipsum}
\usepackage{soul}
\usepackage{indentfirst}
\usepackage{CJK}
\usepackage{flushend}

\begin{document}

\title{\LARGE \bf
	Weakly Supervised Disentangled Representation for Goal-conditioned Reinforcement Learning}

\author{Zhifeng Qian, Mingyu You, Hongjun Zhou and Bin He%
\thanks{Manuscript received September, 9, 2021; Revised November, 29, 2021; Accepted January, 4, 2022. This paper was recommended for publication by Editor K. Jens upon evaluation of the Associate Editor and Reviewers' comments.
This work was partly supported by the National Natural Science Foundation of China under grant no.62073244 and the Shanghai Innovation Action Plan under grant no.20511100500, 20511105802. (Corresponding author: Hongjun Zhou.)}
\thanks{Z. Qian, M. You, H. Zhou and B. He are with the College of Electronic and Information Engineering, Frontiers Science Center for Intelligent Autonomous Systems, Tongji University, Shanghai 201800, China. (e-mail: qianzhifeng@tongji.edu.cn; myyou@tongji.edu.cn; zhouhongjun@tongji.edu.cn; hebin@tongji.edu.cn.)}%
\thanks{Digital Object Identifier (DOI): see top of this page.}
}

\markboth{IEEE Robotics and Automation Letters. Preprint Version. Accepted January, 2022}
{Qian \MakeLowercase{\textit{et al.}}: Weakly Supervised Disentangled Representation for Goal-conditioned Reinforcement Learning} 

%


\maketitle

	\begin{abstract}
	Goal-conditioned reinforcement learning is a crucial yet challenging algorithm which enables agents to achieve multiple user-specified goals when learning a set of skills in a dynamic environment. However, it typically requires millions of the environmental interactions explored by agents, which is sample-inefficient.
	In the paper, we propose a skill learning framework DR-GRL that aims to improve the sample efficiency and policy generalization by combining the Disentangled Representation learning and Goal-conditioned visual Reinforcement Learning. In a weakly supervised manner, we propose a Spatial Transform AutoEncoder (STAE) to learn an interpretable and controllable representation in which different parts correspond to different object attributes (shape, color, position). Due to the high controllability of the representations, STAE can simply recombine and recode the representations to generate unseen goals for agents to practice themselves. The manifold structure of the learned representation maintains consistency with the physical position, which is beneficial for reward calculation. 
	We empirically demonstrate that DR-GRL significantly outperforms the previous methods in sample efficiency and policy generalization. In addition, DR-GRL is also easy to expand to the real robot.
	\end{abstract}

\begin{IEEEkeywords}
Reinforcement Learning, Representation Learning, Deep Learning for Visual Perception.
\end{IEEEkeywords}

\section{Introduction}
%
%
%
%
\IEEEPARstart{R}{einforcement} learning (RL) is a promising algorithm that enables agents to learn a skill by optimizing a specific reward function. Impressive works have been proposed, from playing Atari\cite{Mnih2015HumanlevelCT}, Go \cite{Silver2016MasteringTG,Silver2017MasteringTG} and chess \cite{Schrittwieser2020MasteringAG} to learning robotics manipulation skills \cite{Su2020ReinforcementLB,Yang2020RigidSoftIL,Pedersen2020GraspingUO}.
Standard RL algorithms can only acquire one skill at a time. However, agents are expected to achieve multiple user-specified goals in an unstructured real environment. In this paper, we aim to study goal-conditioned reinforcement learning (\textbf{GRL}) to achieve this by maximizing the average success rate conditioned on diverse goals.

The core issue of goal-conditioned RL is to improve the sample efficiency. Setting multiple goals during training requires agents to explore millions of interactive samples, which is exacerbated by visual observations. To this end, the previous works \cite{Nair2018VisualRL,Srinivas2020CURLCU} are dedicated to learning an expressive low-dimensional representation from high-dimensional observations. 
During training, it's laborious to manually set various visual goals for the agent. Therefore, the representations can also be used to generate new goals to train the agent. 
However, these representations without additional constraints have no explicit physical meaning, and it cannot be explained whether these representations focus on the task-relevant information. Moreover, the goals generated based on these representations become uncontrollable and sometimes ineffective. These goals would confuse the agent, thereby making training efficiency lower and policy performance worse.
If the agent can learn an interpretable and controllable representation, it can imagine reasonable goals on top of it to practice itself, which would improve the learning efficiency.

\begin{figure}[t]
	\centering
	\includegraphics[width=7cm]{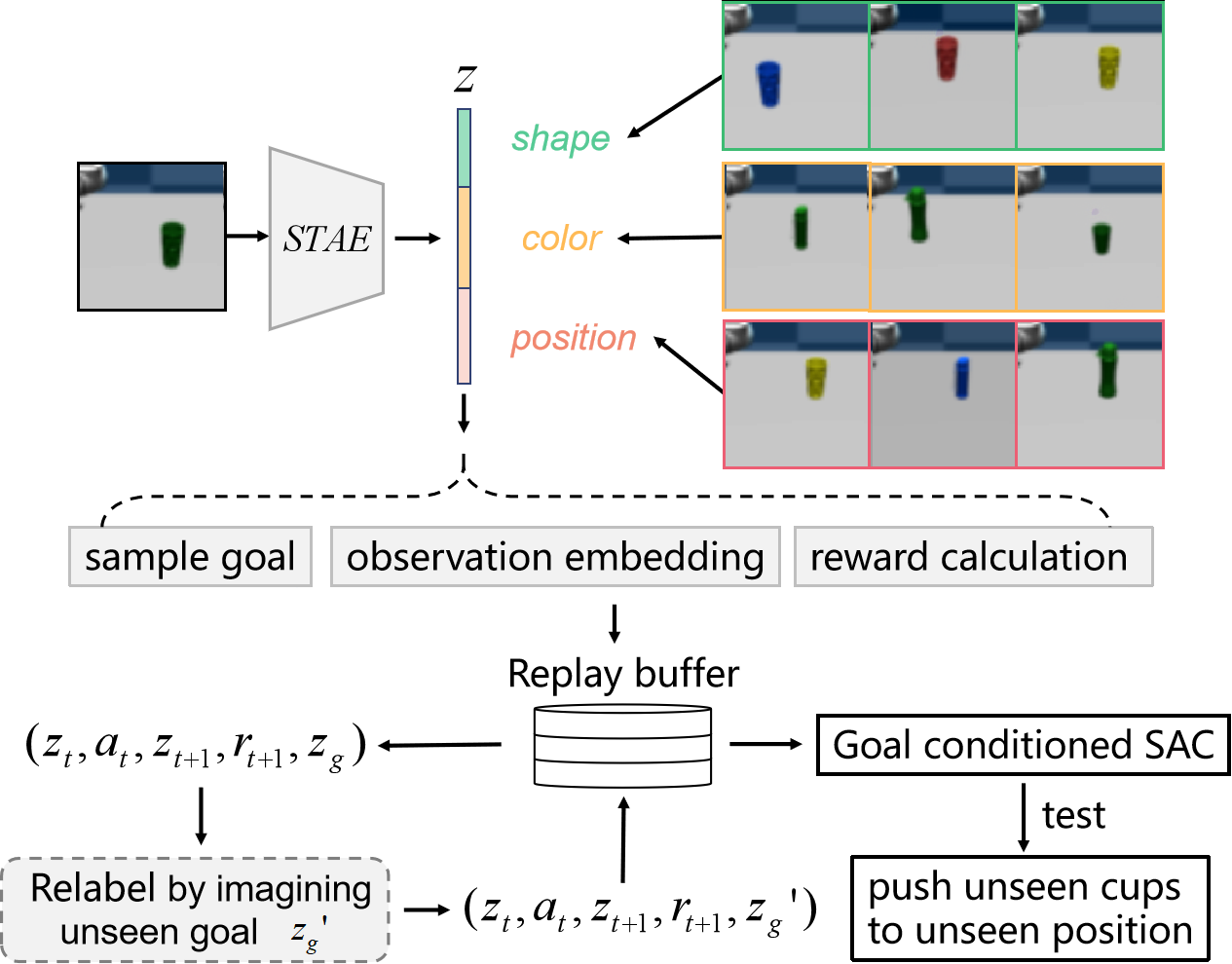}
	\caption{Overview of our framework DR-GRL. First, we train a spatial transform autoencoder (STAE) to learn an object-oriented representation in which different parts correspond to the object attributes. Then we use the learned representation space as the goal space, input space and reward calculation in GRL. Moreover, DR-GRL can generate unseen goals to relabel data in replay buffer to imagine more interactive samples. Finally, we adapt SAC to learn a goal-conditioned policy to push cups in simulation and the real world.}
	\label{whole}
	\vspace{-1.6em}
\end{figure}

We draw inspiration from the disentangled representation learning in computer vision. Disentangled representation learning (\textbf{DRL}) \cite{Feng2018DualSD,ge2021zeroshot, Niemeyer2020GIRAFFE} has developed in recent years, which investigates how to learn the disentangled representations of attributes, e.g. the object shape, color or position. 
However, in this paper, there are two key differences: (i) In RL, the concern is whether the change in the disentangled representation space maintains monotonously consistency with the change of real physical attributes. For example, the distance between representations should increase as the spatial distance of objects increases. (ii) Vanilla methods require samples to provide attribute templates to generate new samples, but we expect to generate goals with attributes out of the position distribution. In addition, they pay more attention to the appearance and texture rather than the position.

To this end, we propose a skill learning framework \textbf{DR-GRL} to learn an object-oriented \textbf{D}isentangled \textbf{R}epresentation in a weakly supervised manner to improve the efficiency and generalization of \textbf{G}oal-conditioned \textbf{RL}. The overall framework DR-GRL is shown in Fig. \ref{whole}. We propose a spatial transform autoencoder (\textbf{STAE}) to learn a representation that disentangles the shape, color, and position of the object without the definite attribute labels. 
The manifold structure of the learned representation maintains consistency with the physical position, which facilitates efficient reward calculation. 
By recombing and recoding the disentangled representation, STAE can generate unseen objects at any position of the image. The corresponding representation can be used as goals for agents to practice and to relabel data in replay buffer. It is worth noting that by aligning the position encoding of the simulation and the real world, DR-GRL can transfer the policy to the real world without fine-tuning. We perform experiments on the pushing cup task in the simulation and the real world. The empirical results demonstrate that DR-GRL significantly improves the efficiency and generalization of the goal-conditioned visual reinforcement learning.

Our main contributions are as follows:
\begin{enumerate}
	\item We propose DR-GRL to learn an interpretable and controllable disentangled representation that enables goal-conditioned visual RL to improve the sample efficiency and policy generalization.
	\item We design STAE to make the disentangled representation consistent with the physical position. By simply recombining and recoding the representation, STAE can generate diverse goals out of the distribution. 
	\item We empirically demonstrate the effectiveness of our DR-GRL in pushing cup task in both simulation and the real world.
\end{enumerate}

\section{Related works}
\subsection{Goal-conditioned Reinforcement Learning}

Many prior works on GRL are devoted to learning goal-conditioned policies that can enable agents to achieve diverse goals. 
To increase sample efficiency, Schaul et al. \cite{Schaul2015UniversalVF} provide a simple mechanism to substitute for the old goals in off-policy settings. Andrychowicz et al. \cite{Andrychowicz2017HindsightER} propose Hindsight Experience Replay (HER) to relabel goals with the future goal spaces. However, these methods are only suitable for the state input, e.g. object coordinates rather than visual input. They can directly change the state as a new goal, which is difficult to implement in visual input. Nair et al. \cite{Nair2018VisualRL} learn a representation by a variational autoencoder (VAE) \cite{Kingma2014AutoEncodingVB} to adapt to visual GRL, which samples latent representation as new goals for training. 
However, vanilla VAE cannot generate valid and feasible goal images.
Later, Nair et al. learn a context-conditioned VAE in \cite{Nair2019ContextualIG} and a Vector Quantised-Variational VAE \cite{Oord2017NeuralDR} in \cite{Khazatsky2021WhatCI} to solve this issue by generating more consistent and reasonable goal images. Wang et al. \cite{Wang2020ROLLVS} learn an object-VAE to encode the object segmentation, which can sample the goal embeddings only related to the target object.
However, these methods cannot flexibly control the object generation to unseen positions, which limits the generalization of the policy. By conveniently recoding the learned encoding, our DR-GRL can generate unseen objects at any position of the image, which is out of the training distribution.

\subsection{Representation Learning for Vision-based control}
For vision-based control tasks, many related works aim to learn an expressive representation to improve the sample efficiency of visual RL. Park et al. \cite{Park2020LearningTA} learn the se(3) constrained features to directly represent the object pose for geometric imitation of object manipulation. However, this method requires certain supervision signals in the learning process. Srinivas et al. \cite{Srinivas2020CURLCU} propose a framework that integrates contrastive learning with model-free RL. 
Pathak et al. \cite{Pathak2017LearningFB} learn visual representations by roughly segmenting objects in an unsupervised manner. Kulkarni et al. \cite{kulkarni2019unsupervised} learn the object keypoints for control from videos without supervision. Zhang et al. \cite{zhang2020learning} learn invariant representations with the properties of bisimulation metrics for effective downstream control. 
However, the above learned representations have ambiguous physical meanings. In contrast, our DR-GRL learns the representations which can disentangle different attributes of the objects in a weakly supervised manner.

\subsection{Disentangled Representation in Computer Vision}
In computer vision, disentangled representation learning has been explored by prior works \cite{Feng2018DualSD, chen2018isolating, zhang2021sill}. Niemeyer et al. \cite{niemeyer2021giraffe} propose a novel method for generating scenes in a controllable and photorealistic manner. Yunhao et al. \cite{ge2020zero} disentangle different attributes into a representation by swapping operations to synthesize new samples. However, the disentangled representation in computer vision research field cannot be consistent with the physical positions of objects in the images. Our DR-GRL introduces a spatial transform module to learn the representations maintaining the manifold structure of the physical positions, which is empirically demonstrated in \ref{exp-consist}.

\begin{figure}[t]
	\centering
	\includegraphics[width=8cm]{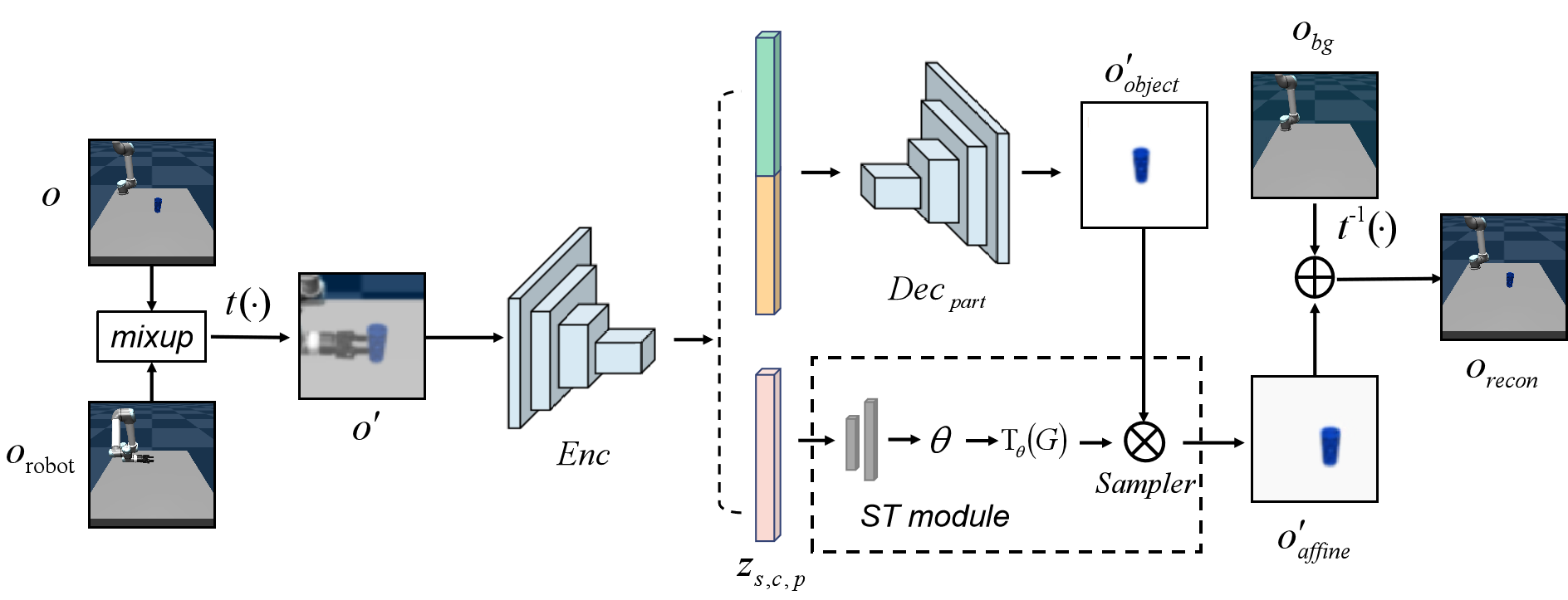}
	\caption{Architecture of spatial transform autoencoder(STAE). }

	\label{network}
	\vspace{-1.6em}
\end{figure}

\section{method}
DR-GRL is a skill learning framework for combining disentangled representation learning with the goal-condition RL. In DR-GRL, we design a spatial transform autoencoder (\textbf{STAE}) to learn interpretable representations to disentangle the latent space in a weakly supervised manner. In the following, we start by introducing the architecture and components of STAE, then show our weakly supervised progressive swap strategy of STAE for representation learning, and finally describe how the representation is applied to goal-conditioned RL.

\subsection{Architecture of Spatial Transform Autoencoder}\label{sec-net}

\begin{figure}[t]
	\centering
	\includegraphics[width=6cm]{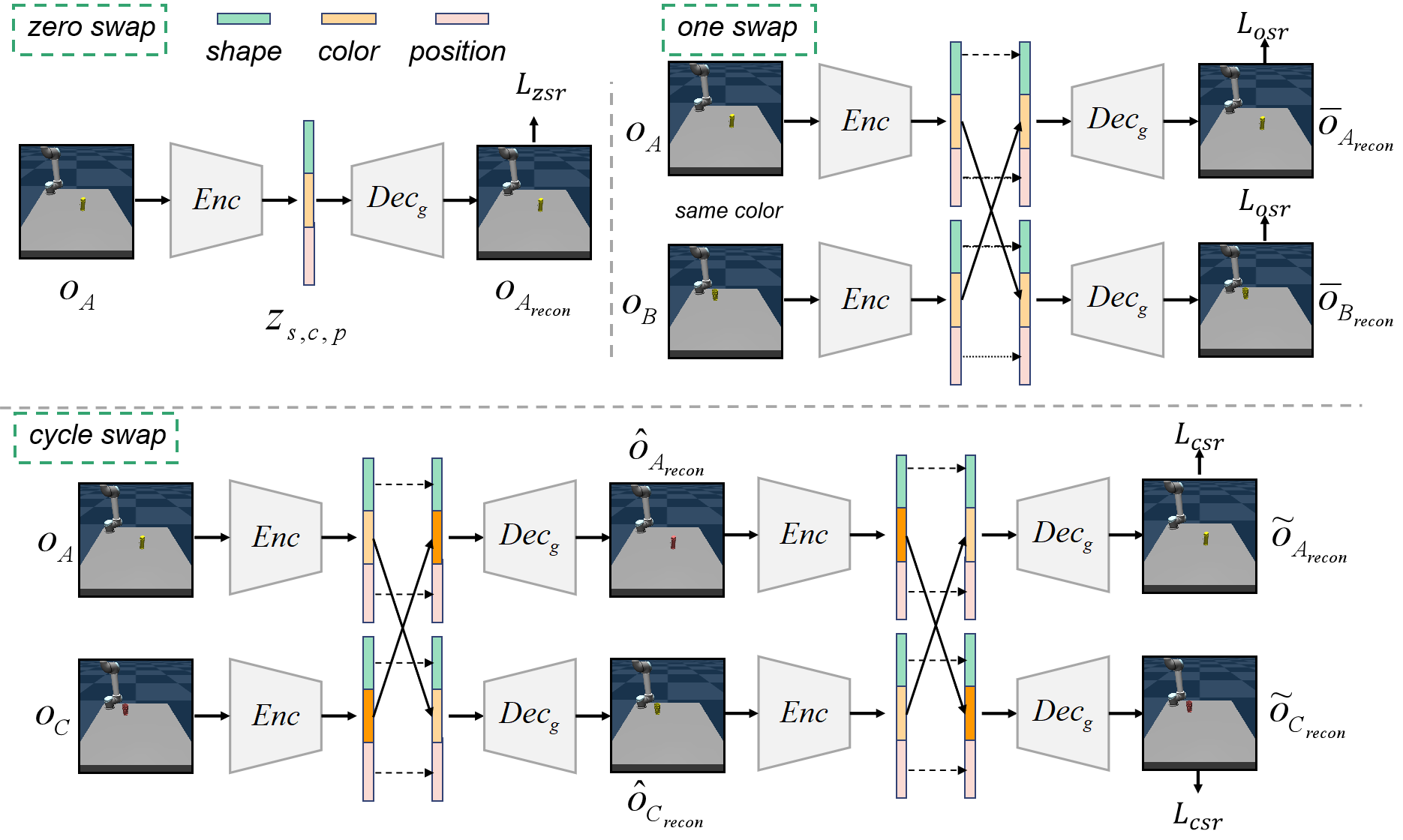}
	\vspace{-1em}
	\caption{Schematic diagram of the weakly supervised progressive swap strategy. There are three stages including zero swap, one swap and cycle swap. All these operations only use the reconstruction loss to optimize STAE. The core of the progressive swap strategy is to swap representations and reconstruct the original image, thereby disentangling the encoding of different attributes.}
	
	\label{swap}
	\vspace{-1.8em}
\end{figure}

The goal of STAE is to take weakly labeled samples as input and train an autoencoder which achieves controllable disentangling in the latent space. The architecture of STAE is shown in Fig. \ref{network}. Give a source observation $o \in \mathbb{R}^{C \times H \times W}$ and a robot observation in random pose  $o_{robot} \in \mathbb{R}^{C \times H \times W}$, we perform two steps of data augmentation on it, as is described as follow:
\begin{equation}\label{data-aug}
	\begin{split}
		o' =  t(m(o, o_{robot})) = t(\lambda_{mixup}  \cdot o + (1 - \lambda_{mixup}) \cdot o_{robot}) 
	\end{split}
\end{equation}
where $m$ means the mixup operation \cite{zhang2018mixup} and $t$ means other common data augmentations such as CenterCrop and Resize. $\lambda_{mixup}$ is a parameter to balance the weight of the two images. The mixup operation $m$ is designed to eliminate the interference of the robot pose on the disentangled representation, since the representation focuses on effective information about the object.
Then, we input the mixed image $o' \in \mathbb{R}^{C \times H' \times W'}$ to the encoder $Enc$ to extract various attributes of the object into the latent representations $z$. In our setting, we wish to disentangle the latent representations $z_{s,c,p}$ related to three attributes of the shape, color and position of the cup. 

Different from the general disentangled representation learning methods, robot tasks pay more attention to the position attributes of objects. We want position encodings to be highly consistent with the real position of objects, rather than the abstract encodings.
So the latent representations here are divided into two parallel branches.  
$z_{s,c}$ of the shape and color attributes are input to the decoder $Dec_{part}$ to reconstruct the cup $o'_{object} \in \mathbb{R}^{C \times H' \times W'}$. In particular, here we generate the cup in the middle of the table, focusing only on the shape and color rather than the position. 

In the other branch, the spatial transform (\textbf{ST}) module takes the position representation $z_{p}$ as input and predicts the translation parameters to translate $o'_{object}$ to $o'_{affine} \in \mathbb{R}^{C \times H' \times W'}$. Finally, we perform an inverse data augmentation operation on $o'_{affine}$ and add it to a basic background image $o_{bg} \in \mathbb{R}^{C \times H \times W}$ to obtain the final reconstructed image $o'_{recon}$, which is described as follow:
\begin{equation}\label{data-add}
	o_{recon} = t^{-1}(o'_{affine}) + o_{bg}
\end{equation}
where $o_{bg}$ can be easily obtained in many ways, such as using Gaussian Mixture-based Background Segmentation algorithm \cite{Wang2020ROLLVS, Zivkovic2006EfficientAD} implemented in OpenCV \cite{bradski2008learning}. In this paper, we simply record $o_{bg}$ before the task is performed.

The ST module is introduced to allow CNNs to perform affine transformation without additional supervision signal inspired by \cite{10.5555/2969442.2969465}. The affine transformation can be expressed by the following matrix operation:
\begin{equation}\label{affine}
	\left(
	\begin{array}{c}
		x'\\
		y'
	\end{array}
	\right) = 
	\begin{bmatrix}
		\vartheta_1&\vartheta_2&\vartheta_3\\
		\vartheta_4&\vartheta_5&\vartheta_6
	\end{bmatrix}
	\left(
	\begin{array}{c}
		x\\
		y\\
		1
	\end{array}
	\right)
\end{equation}
where $\vartheta$ is the affine transformation matrix parameter. $x'$ and $y'$ are pixel coordinates in the original 2D plane while $x$ and $y$ are pixel coordinates after transformation. Unlike \cite{10.5555/2969442.2969465}, we set the first two columns of the matrix as the identity matrix and only predict a two-dimensional vector $[\vartheta_3, \vartheta_6]$ to realize translation transformation. Such a specific design aims to give stronger semantic information to the position representation in the process of reconstructing images, which is highly relevant to the robot task. In addition, interpretable position representation can help improve the controllability and generalization of goal sampling in RL, which will be described in Section \ref{sec-rl}.

\subsection{Weakly Supervised Progressive Swap Strategy for Disentangled Representation Learning}\label{sec-learn}

While training our STAE, we implement a weakly supervised progressive swap strategy to learn the disentangled representations. Weak supervision means we only know which attributes are the same between different samples without the definite attribute labels. The schematic diagram of the weakly supervised progressive swap strategy is shown in Fig. \ref{swap}. 

There are three stages in our learning strategy, which are zero swap, one swap and cycle swap. In the zero swap stage, the encoder $Enc$ first encodes $o_A$ to the representations $z_{s,c,p} = Enc(o_A) = [z_s, z_c, z_p]$, and then the global decoder $Dec_g$ decodes to reconstruct the original input. We wish the representations $z_{s,c,p}$ can contain as much information of the original image as possible. The objective of optimization here, just like common autoencoder, is zero swap reconstruction loss $L_{zsr} = ||o_A - o_{A_{recon}}||_1$.

In the one swap stage, given a pair of inputs $(o_A, o_B)$ with one same attribute, e.g. the color, the encoder encodes them to the representations. If the representations are disentangled, the different segments of the representations are wished to represent different object attributes. When two samples share the same color attribute, we swap each other's color encodings $z_c$ and keep other attribute encodings unchanged. If the swap attribute encodings are disentangled enough and do not contain the information of other attributes, the decoder can still reconstruct their original images. We perform one swap operation for all three attributes. The objective is the one swap loss, which is denoted as 
\begin{equation}\label{l-osr}
	L_{osr} = ||o_A - \overline{o}_{A_{recon}}||_1 + ||o_B - \overline{o}_{B_{recon}}||_1
\end{equation} 

In the cycle swap stage, we repeat the operation of the previous stage twice. We encode two images, swap encodings, decode, encode again, swap again to restore the original feature and decode again. The difference is that the objects in two input images do not need to have the same attributes. Without additional supervision on the intermediate images $ \hat{o}_{recon}$, we only train the network by the cycle swap reconstruction loss $L_{csr}$ of the final generated images, which is denoted as
\begin{equation}\label{l-csr}
	L_{osr} = ||o_A - \hat{o}_{A_{recon}}||_1 + ||o_C - \hat{o}_{C_{recon}}||_1
\end{equation} 
The cycle swap loss enforces our STAE to disentangle the different semantics of different parts on the latent representations. It ensures that the representations of one attribute do not contain information about other attributes.

Zero swap operation ensures the representation containing enough information of the original images. One swap operation can disentangle the shape and color attributes of the objects, while the cycle swap operation can further improve the disentangling effect.
Through the weakly supervised progressive swap strategy, STAE can disentangle the representations corresponding to different object attributes. In addition, by recombining shape and color representations and recoding the position representation, STAE can render an unseen cup to any position of the image, which can be used to sample diverse goals for the agent to practice itself.
\vspace{-0.7em}

\subsection{Goal-conditioned RL with Learned Representations}\label{sec-rl}
Goal-conditioned RL aims to train a policy that can achieve a set of goals. In principle, Goal-conditioned RL can model the variable goal $g$ into any standard reinforcement learning algorithm. We choose SAC which is an effective off-policy RL algorithm. 
With our modification, SAC learns a policy $\pi_{\varphi}$ and two Q-value critics $Q_{\phi_1}$, $Q_{\phi_2}$ with respect to a goal distribution $g \sim G$. We input the learned position representation as observation embeddings into $\pi_{\varphi}$. The goal-conditioned critics $Q_{\phi_i}$ are learned by optimizing the Bellman error in Equ. \ref{l-Q} while the goal-conditioned policy $\pi_{\varphi}$ is optimized by the expected return of its action in Equ. \ref{l-pi}
\begin{align}
	L(\phi_i&, R)= \mathbb{E}_{d\sim R}[(Q_{\phi_i}(o,a,g)-(r+\gamma(1-d)T))] \label{l-Q}
	\vspace{1.4em}
	\\ &T = \min_{i=1,2}Q_{\phi_i}^*(o',a',g') - \alpha log \pi_{\varphi}(a'|o') \label{target}
\end{align}
\begin{equation}\label{l-pi}
	L_\varphi = \mathbb{E}_{a\sim \pi}[Q_{\phi_i}(o,a,g)-\alpha log \pi_\varphi(a|o,g)]
\end{equation}
where $d = (o,a,o',g,r,d)$ is a tuple with observation $o$, action $a$, observation at the next moment $o'$, goal $g$, reward $r$ and done signal $d$ in the replay buffer $R$. 
$\alpha$ is a positive entropy coefficient to balance the value function and entropy maximization. 

\subsubsection{Reward Specification}
We define a goal-conditioned reward function $r(o,g)$ for the vision-based RL. Due to the real-time lighting changes in the real scene, the Euclidean distance in the pixel space may cause a large variance in the reward. All we care about is whether the object reaches the goal position $z_p^g$, which can be reflected in the learned disentangled representation. Therefore, we use the Euclidean distance of the position representations to calculate the reward, which is denoted as
\begin{equation}
	\label{reward}
	r(o,g) = -||Enc(o)_p - z_{p}^g||_1
\end{equation}

In particular, the distance in our disentangled representation space is consistent with the real position of the object. This consistency cannot be guaranteed in the representation space learned by vanilla autoencoders since there are no additional constraints on representation in training. We empirically show that taking our disentangled representation distance as a reward can significantly improve the training efficiency of RL.

\begin{algorithm}[t]
	\caption{DR-GRL: disentangled representation for goal-conditioned RL}
	\label{algorithm_train}
	\begin{algorithmic}[1]
		\State Collect Data with different attributes of objects.
		\State Train STAE on Data by progressive swap strategy (\ref{sec-learn})
		\State Initialize goal-conditioned SAC Algorithm.
		\State \textbf{for} i = 1:E episodes do
		\State \quad Sample goals $z_g \sim U(z_g)$ based on the representations.
		\State \quad \textbf{for} step = 1:N do
		\State   \quad \quad Sample $a_t \sim \pi(o,z_g)$ to interact.
		\State   \quad \quad Calculate the reward $r_t$ according to Equ. \ref{reward}.
		\State   \quad \quad Store tuple $(o_t, a_t, o_{t+1}, r_t, z_g, d)$ in replay buffer $R$.
		\State   \quad \quad \textbf{for} k = 1:K do
		\State   \quad \quad \quad Sample tuples $(o_t, a_t, o_{t+1}, r_t, z_g, d)$ in $R$.
		\State   \quad \quad \quad Sample $z'_g \sim U(z_g)$ to replace the $z_g$ in the tuples.
		\State   \quad \quad \quad Calculate new reward $r'_t$ according to Equ. \ref{reward}.
		\State 	 \quad \quad \quad Store relabeled tuples in $R$. 
		\State   \quad \quad \textbf{end for}
		\State 	 \quad \quad perform one step of optimization using Equ.\ref{l-Q}, \ref{l-pi}
		\State  \quad \textbf{end for}			
		\State \textbf{end for}		
		\State\textbf{return} {$\theta$}
	\end{algorithmic}
\end{algorithm}

\subsubsection{Goal Generation and Relabeling}  We generate various goals to train the policy $\pi$ during training. For vision-based settings, it is not convenient to collect specific images as goals in both simulation and the real world. The previous method \cite{Nair2018VisualRL} generates goals by sampling the latent features of VAE, which lacks semantic consistency and generalization of new positions. Our STAE can achieve zero-shot goal generation in a highly controllable manner. STAE recombines the shape and color representations to render new objects. Furthermore, STAE can recode the position representation by randomly sampling within the specified range to generate objects to any desired position of the images. And the interpretability ensures the validity of the generated goals.

For a tuple $d = (o,a,o',g,r)$ in the replay buffer $R$, we can relabel the tuple with the generated goal $g'$ and reward function $r'(o, g)$. Therefore, more tuples like $d' = (o,a,o',g',r')$ can be obtained without the time-consuming rollouts.  
Previous methods \cite{Nair2018VisualRL, Andrychowicz2017HindsightER} are proposed to sample goals from a trajectory or from learned priors, which brings bias to the goal distribution. We use highly controllable disentangled representations to generate goals under uniform distribution $g' \sim U(g)$ in the position representation space. The empirical results demonstrate that the goals generated from the disentangled representations greatly improve the sample efficiency.

\subsubsection{Algorithm Summary}
We summarize the whole algorithm of disentangled representations for goal-conditioned reinforcement learning (DR-GRL) in Algorithm. \ref{algorithm_train}. We first collect data including different attributes of objects. Then, by the weakly supervised progressive swap strategy, STAE is trained on the data to learn a disentangled representation. We use the obtained disentangled representation space for the goal space, observation embedding and reward calculation in GRL. For every tuple in replay buffer, we can relabel the tuple by sampling new goals and calculating new rewards using Equ. \ref{reward}. At last, we use the adapted SAC \cite{haarnoja2018soft} to learn a goal-conditioned policy and critics by Equ.\ref{l-Q}, \ref{l-pi}.

\section{experiments}\label{experiments}
To verify the effectiveness of our DR-GRL, we perform experiments on the pushing cup task by simulation and real robotic arms respectively. We train our DR-GRL on 4 NVIDIA TITAN X GPUs. We build a Mujoco simulation environment, named  \emph{UR Pusher}, which must push the cup to a target position. In the real world, we perform experiments on a UR5 robotic arm and a 2F Robotiq gripper.
Our experiments address the following questions:
(1) Does STAE in DR-GRL disentangle the object attributes and generate unseen images as goals for GRL? (2) How does our DR-GRL compare with other methods in improving the sample efficiency and performance of GRL? (3) Can the disentangled representation maintain consistency with the physical position? (4) Is each component of our DR-GRL really effective for skill learning? (5) Can DR-GRL conveniently realize the policy transfer from sim to real?

\begin{figure}[t]
	\centering
	\includegraphics[width=5.5cm]{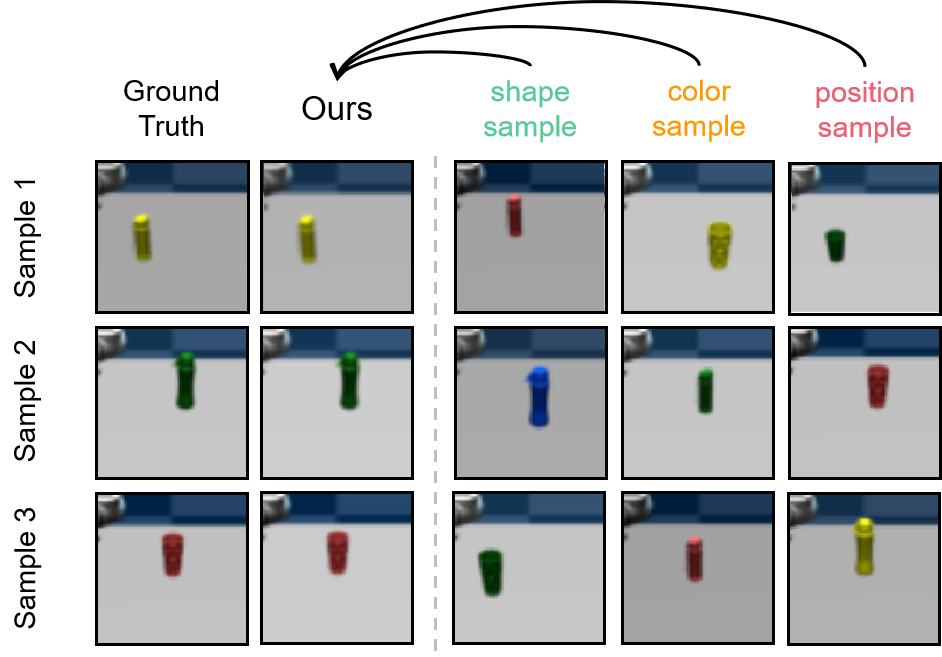}
	\vspace{-0.5em}
	\caption{Here are new samples generated by STAE based on the three attribute examples. Each row shows a generated sample. The first two columns are the ground truth and the generated samples. The three columns on the right are attribute samples to provide different attribute representations.} 
	\label{pic-generation}
	\vspace{-1.5em}
\end{figure}
\begin{figure}[!h]
	\centering
	\includegraphics[width=5.5cm, height=3.cm]{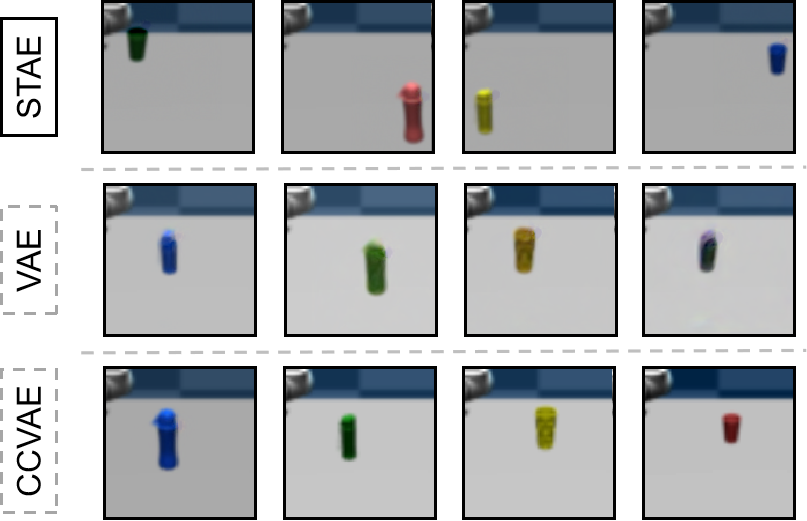}
	\caption{Here shows some examples of new goals generated by our STAE and VAE. Note that the position distribution of the cups in our training set is about a quarter of the center of the image.} 
	\label{pic-generation2}
	\vspace{-1.6em}
\end{figure}

\subsection{DR-GRL architecture.}\label{exp-archi} 
For all experiments, we modify the U-net \cite{10.1007/978-3-319-24574-4_28} by introducing the ST module from \cite{10.5555/2969442.2969465} as our STAE. We add 2 fully connected layers in the middle to obtain the disentangled representation. We also add the spatial attention layers in CBAM \cite{2018CBAM} to increase the expressiveness of the network. The representation dimension is 100, in which the shape and color dimension are both 49, and the position dimension is 2. 

In principle, any RL algorithm can be built on top of the learned representation space. We modify Soft Actor Critic(SAC) \cite{haarnoja2018soft} to adapt to the GRL setting. The input space of the policy is 4-dimensional including the goal encoding and the current observation encoding. The output space includes the 6-dimensional tool center position of the UR5 robotic arm and the 1-dimensional switch state of the gripper.

\subsection{Disentangled Representation and Goal Generation}\label{exp-generation}
\begin{figure}[h]
	\vspace{-1.5em}
	\centering
	\includegraphics[width=6cm]{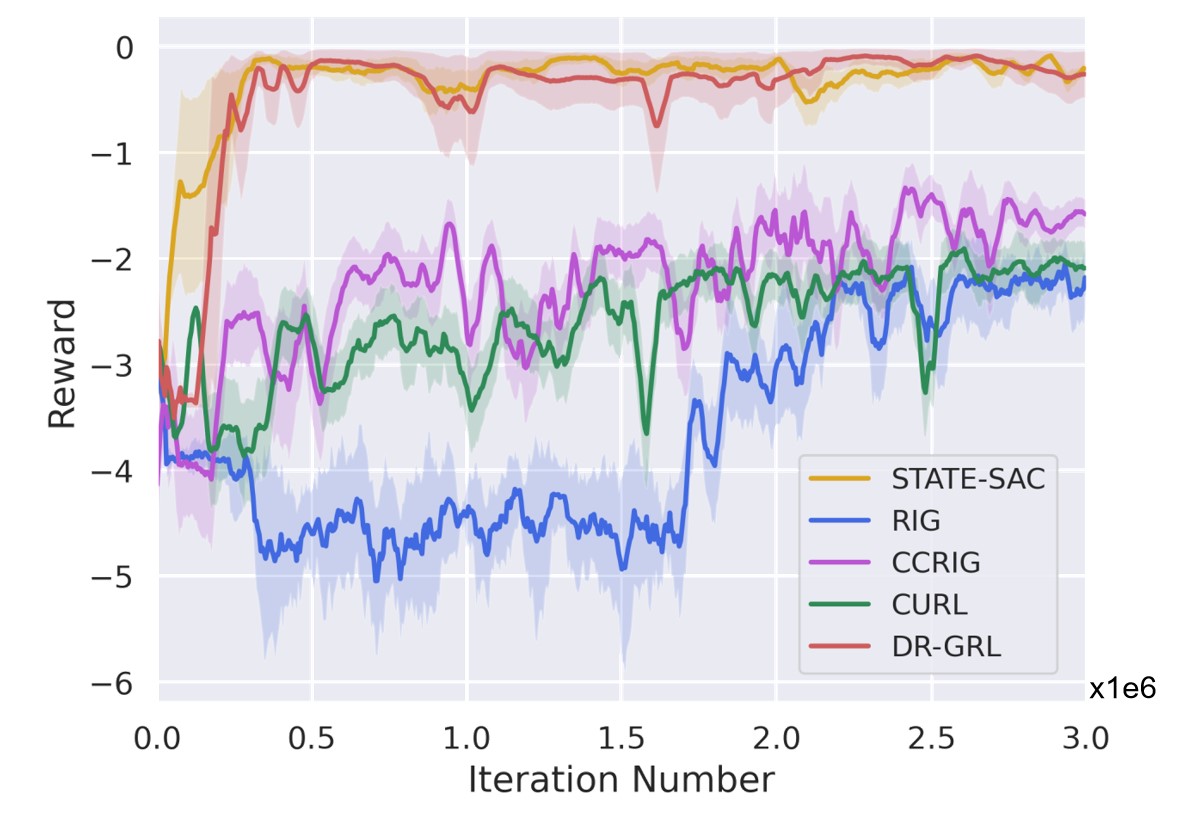}
	\vspace{-1em}
	\caption{Simulation results. DR-GRL outperforms other state-of-art methods in both sample efficiency and performance, except for STATE-SAC which uses ground truth state for input. Each method is evaluated with 5 seeds in \emph{UR Pusher} environment.}
	\label{pic-reward}
\end{figure}

\begin{figure}[t]
	\centering
	\includegraphics[width=6cm]{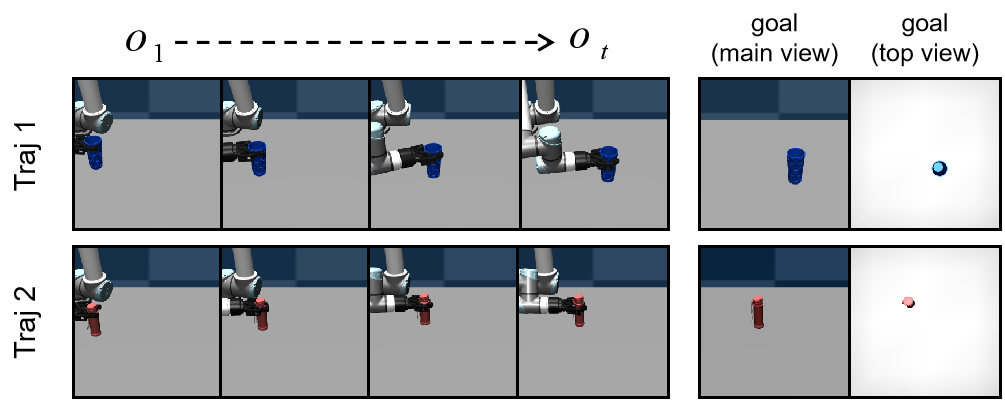}
	\vspace{-1em}
	\caption{Some trajectories to successfully achieve the sampled goals in simulation. Each row shows a robot execution trajectory under a given target. }
	\label{pic-traj}
	\vspace{-0.6em}
\end{figure}

\begin{figure}[h]
	\centering
	\includegraphics[width=7.5cm]{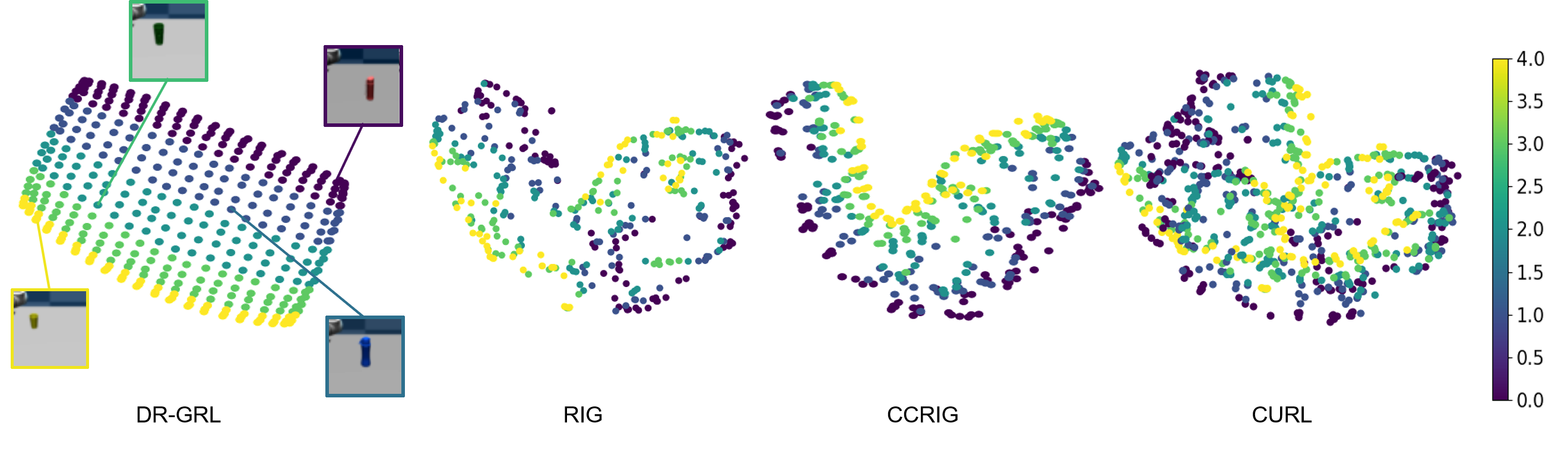}
	\caption{Visualization of the representations learned by different methods. The dimension is reduced to 2 using t-SNE. We sample 20*20 positions of the cups and use colors to indicate the changes. The color scale of 0 or 4 indicates that the position is in the upper left or lower right corner of the table. }
	\label{pic-stne}
	\vspace{-0.52cm}
\end{figure}

To verify the disentangling of the learned representations, we train our STAE by the dataset on cups with different attributes collected in the \emph{UR Pusher} simulation. There are 4 shapes, 4 colors and 16 positions for the cup under 5 different lighting conditions in the dataset. The qualitative results are given in Fig. \ref{pic-generation}. As is shown, given the example of each attribute, our STAE can combine the specific attribute to synthesize a new and clear image, which means that the learned representation only contains information about the corresponding attribute.

We expect to control disentangled representation to generate unseen goals for GRL. 
We compare STAE with other generative models, including VAE \cite{Kingma2014AutoEncodingVB} used in RIG\cite{Nair2018VisualRL} and CCVAE used in CCRIG\cite{Nair2019ContextualIG}. As is shown in Fig. \ref{pic-generation2}, VAE can generate some blurred cups, while CCVAE can generate more clearly. However, both of them can only generate cups in the positions that have appeared in the training data.
For comparison, by recoding the position encoding, STAE can render the cup to the position out of the training distribution, even anywhere in the images. The highly controllable generation allows goal-conditioned agents to imagine diverse goals during training, which can improve the generalization of the policy.

\subsection{Simulation experiments of pushing cups}\label{exp-rl}

To answer the second question, we compare DR-GRL with previous methods in \emph{UR Pusher}. RIG: \cite{Nair2018VisualRL} trains a VAE to learn a latent representation for visual GRL. CURL: \cite{Srinivas2020CURLCU} combines RL with representation learning to improve the sample efficiency. 
CCRIG:  \cite{Nair2019ContextualIG} trains a CCVAE to generate more reasonable goal images for visual GRL.
STATE-SAC: vanilla SAC \cite{haarnoja2018soft} with access to object coordinates, which is used as the upper bound of visual RL algorithms. Since CURL\cite{Srinivas2020CURLCU} and STATE-SAC\cite{haarnoja2018soft} are only trained to achieve a single goal, we adapt them to diverse goals by modifying the goal-conditioned actors and critics. In addition, for these methods can't generate new goals, we artificially set some goals from the same goal distribution in their representation space during training. For a fair comparison, we use SAC as the same RL algorithm to train DR-GRL, RIG, CURL, CCRIG and STATE-SAC.

The simulation results are shown in Fig. \ref{pic-reward}. We normalize the reward on top of the representation spaces learned by different methods. We define success within 1cm from the goal position, and normalize the representation distance between the success boundary and the goal to -1.
The results demonstrate that compared with RIG, CCRIG and CURL, our DR-GRL converges faster and significantly achieves a better performance. In addition, our DR-GRL is more stable during training.
We suspect that used for observation embedding and reward calculation, the disentangled representation can focus on task-relevant information, thereby improving the convergence and stability of DR-GRL. Furthermore, without access to the object state, our DR-GRL achieves almost the same performance as STATE-SAC.

\begin{table}[H]\footnotesize
	\caption{Quantitative results in simulation}
	\label{table1}
	\begin{center}
		\linespread{1.2}\selectfont
		\begin{tabular}{m{2cm}<{\centering}|m{2.7cm}<{\centering}|m{2.78cm}<{\centering}}
			
			\Xhline{1.5pt}
			
			\multirow{2}{*}{Methods} & \multicolumn{2}{c}{Success rate(\%) / Reward}  \\
			\cline{2-3}
			& seen cups & unseen cups\\
			
			\Xhline{1.1pt}
			STATE-SAC\cite{haarnoja2018soft} (as upper bound) &89.33 / -0.18 $\pm$ 0.11 & 89.33 / -0.18 $\pm$ 0.11 \\
			\hline
			RIG\cite{Nair2018VisualRL} & 16.00 / -2.47 $\pm$ 0.32 & 5.33 / -4.62 $\pm$ 0.25\\
			\hline
			CCRIG\cite{Nair2019ContextualIG} & 28.67 / -1.63 $\pm$ 0.24 & 25.33 / -1.81 $\pm$ 0.25\\
			\hline
			CURL\cite{Srinivas2020CURLCU}  & 20.67 / -2.03 $\pm$ 0.29 & 10.67 / -4.55 $\pm$ 0.27\\
			\hline
			\textbf{DR-GRL}	  & \textbf{85.33 / -0.21 $\pm$ 0.19} & \textbf{84.00 / -0.28 $\pm$ 0.21} \\
			\Xhline{1.5pt}
		\end{tabular}
	\end{center}
	\vspace{-0.4cm}
\end{table}

During testing, each method samples 30 goals from the same distribution and tests 5 times. Some successful trajectories in simulation are pictured in Fig. \ref{pic-traj}. We evaluate the performance of each method by comparing the average success rate and reward on the seen cups and unseen cups respectively, which is shown in Table. \ref{table1}. The seen cups and unseen cups vary in shape and color. And the positions of cups are sampled from the uniform distribution.
We notice that both RIG and CURL can only push the seen cups to several goals, and can hardly push the unseen cups. We suspect that the representations learned by these methods cannot disentangle the shape of the cup from the entire image, so they have no generalization ability to the unseen object. CCRIG can push seen and unseen cups with similar success rates, but struggles to achieve a wider range of goals. we consider that it's not possible for CCRIG to train agents by sampling goals whose positions are out of the training distribution.
In addition, the inconsistency problem of other methods also makes the learning more difficult, which illustrates in \ref{exp-consist}. We can see that DR-GRL can push seen and unseen cups with the success rate of 85.33\% and 84.00\%, which is more than \textbf{3x} higher than RIG, CCRIG and CURL. 
Even if STATE-SAC has access to the coordinate states of the cup and the target, our DR-GRL can still achieve particularly close performance.

\subsection{Consistency of Disentangled Representation and Physical Position}\label{exp-consist}

\begin{figure}[b]
	\vspace{-1.65em}
	\centering
	\includegraphics[width=4cm]{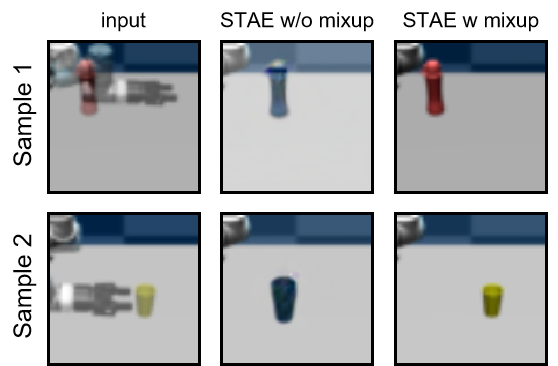}
	\caption{Some reconstruction results through STAE with or without the mixup operation. The results demonstrate that our STAE without the mixup operation cannot clearly reconstruct the task-relevant images.}
	\label{pic-mixup}
	\vspace{-1.5em}
\end{figure}

To answer the third question, we visualize the representations learned by different methods in Fig. \ref{pic-stne}. The dimension is reduced to 2 using t-SNE \cite{Maaten2008VisualizingDU}. We use dots to indicate the cup samples at 20*20 positions, and use colors to indicate the position changes.
As is shown, the representations of RIG and CURL are rather messy, which is unable to distinguish the samples with different physical positions. When the distance between the cup position and the goal becomes larger, the reward calculated in the feature space of RIG and CURL may become smaller, which makes the agent confused and learn sub-optimal actions. 
The representations of CCRIG are relatively better, but the representation distance and physical positions are not monotonically changing.

By contrast, regardless of whether the physical position changes horizontally or vertically, our position representations can well maintain the manifold structure of the original samples. When used for reward calculation, the distance in the disentangled representation space of DR-GRL can directly reflect the actual physical distance between the current state and the goal, which significantly improves the sample efficiency of the GRL. 

\subsection{Ablation Experiments}

\begin{table}[h]
	\vspace{-1em}
	\caption{Ablation Experiments for Swap Strategy}
	\vspace{-1em}
	\label{table_swap}
	\begin{center}
		\begin{tabular}{l|m{0.7cm}|m{0.7cm}|m{3.5cm}}
			\Xhline{1.5pt}
			Model & PSNR & LPIPS & Success rate(\%)/Reward\\
			\hline
			STAE w/o OS & 12.12 & 0.31 & 21.33 / -1.83 $\pm$ 0.35 \\
			\hline
			STAE w/o CS & 29.61 & 0.16 &  72.67 / -0.51 $\pm$ 0.29 \\
			\hline
			\textbf{STAE} & \textbf{34.76} & \textbf{0.028} &  \textbf{85.33 / -0.21 $\pm$ 0.19} \\
			\Xhline{1.5pt}
		\end{tabular}
	\end{center}
	\vspace{-1.5em}
\end{table}	

To answer the fourth question, we train several ablation STAEs to evaluate the contribution of each swap strategy, which is shown in Table \ref{table_swap}. The swap strategies include one swap operation(OS) and cycle swap operation(CS). We compute the distance between ground truth and the generated images based on three attribute examples. We use Peak Signal to Noise Ratio (PSNR) to measure	the image quality and use Learned Perceptual Image Patch Similarity (LPIPS) \cite{Zhang2018TheUE} to indicate the perceptual difference. The larger PSNR and the smaller LPIPS indicate the better image quality. We also use the success rate to evaluate different models.

As is shown, STAE w/o OS struggles to generate the correct image and STAE w/o CS performs a bit better. Our full STAE with OS and CS has the highest PSNR of 34.76 and minimum LPIPS of 0.028. With the improvement of disentangling effect, the performance of the policy also gets better. We consider that OS helps disentangle the shape and color of the objects and CS can further help to disentangle attributes, which can improve the performance of the policy.

\begin{table}[h]
	
	\caption{Ablation Experiments for each component}
	\label{table3}
	\vspace{-0.45cm}
	\begin{center}
		\begin{tabular}{l|c}
			\Xhline{1.5pt}
			Model & Success rate(\%) / Reward \\
			\hline
			DR-GRL w/o M  & 0.00 / -4.53 $\pm$ 0.37       \\
			\hline
			DR-GRL w/o STm,UR  & 18.67  / -1.88 $\pm$ 0.32           \\
			\hline
			DR-GRL w/o UR &  75.33 / -0.37 $\pm$ 0.23             \\
			\hline
			\textbf{DR-GRL}& \textbf{85.33 / -0.21 $\pm$ 0.19}        \\
			
			\Xhline{1.5pt}
		\end{tabular}
	\end{center}
	\vspace{-1.3em}
\end{table}

We also conduct several experiments to analyze the influence of each component in our DR-GRL, which is shown in Table \ref{table3}. The components of our DR-GRL include mixup (M), spatial transform module (STm) and uniform relabeling (UR). We add the checked component in turn to train the models and test them to push seen cups 30 times with 5 seeds.

As we can see, DR-GRL w/o M can't complete the task since the different poses of the robot may seriously affect the disentangled representations without the mixup operation. We show the images after mixup operation and reconstructed images in Fig. \ref{pic-mixup}. We mixup robot images with different poses into the original images. The results show our STAR can clearly reconstruct the task-relevant cup images without being disturbed by the robot.

Since uniform relabeling is uniformly sampled in the representation space, which is learned through ST module, the model cannot remove ST module and retain the UR operation. DR-GRL w/o STm and UR achieves a success rate of 18.67$\%$. Although the model can achieve several goals, the learned representation is inconsistent with the physical position. We believe that the distance in this representation space cannot represent the completion degree of the task.

As is shown in Table \ref{table3}, DR-GRL w/o UR significantly improves the success rate. We consider that the introduction of ST module makes the disentangled representation consistent with the state of the cups. In addition, uniform relabeling can further improve the performance of our DR-GRL since agents can imagine and explore diverse possibilities during training by uniform relabeling.
\vspace{-0.5em}

\begin{figure}[t]
	\centering
	\includegraphics[width=8cm]{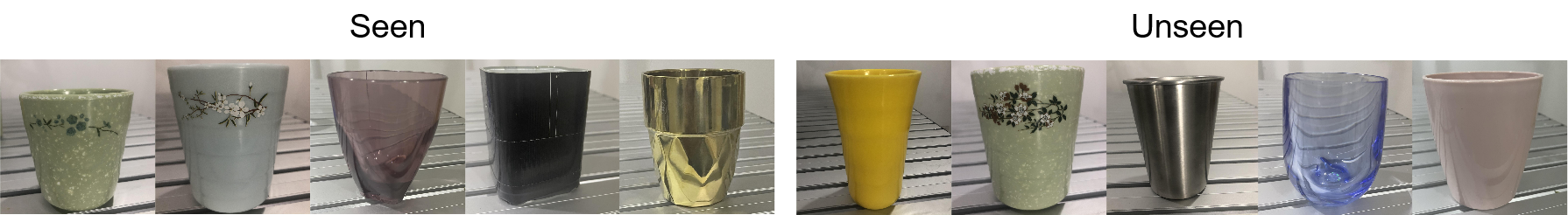}
	\vspace{-0.8em}
	\caption{Diagram of seen cups and unseen cups.}
	\label{cups}
	\vspace{-0.5em}
\end{figure}

\begin{figure}[t]
	\centering
	\includegraphics[width=5.5cm]{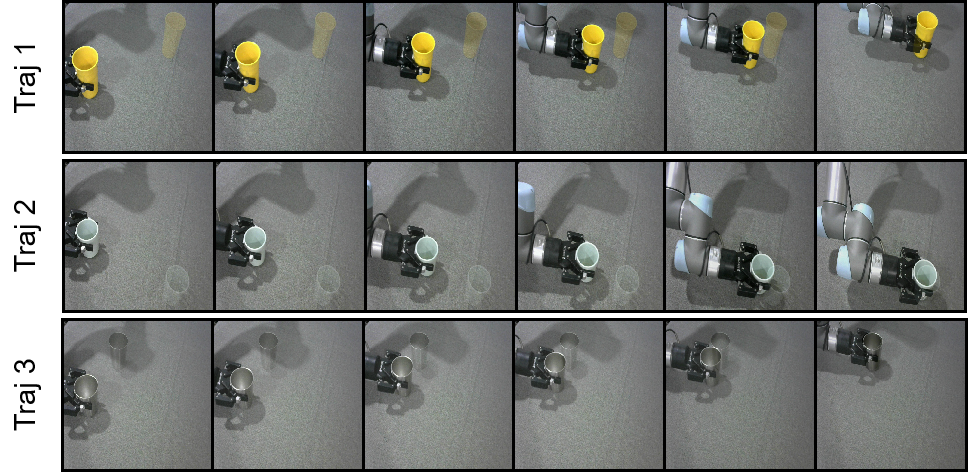}
	\vspace{-0.5em}
	\caption{Some trajectories to successfully achieve the sampled goals in the real world. Each row is a trajectory of the robot pushing the cup to the given goal.}
	
	\label{pic-traj-real}
	\vspace{-1.8em}
\end{figure}

\subsection{Real-world Vision-based Manipulation}
DR-GRL can easily transfer the control policies from simulation to the real world by separating perception and control. 
In our DR-GRL, STAE focuses on the visual deviation of the relative position of the object and the center of the console. By simply mapping the position representation, the policy learned in simulation can be directly deployed on real robots without fine-tuning.

We train the STAE of DR-GRL with 90 real images, including 5 cups in 9 positions under 2 lighting conditions. Then we perform a simple linear mapping on the position representation of the simulation and real images. We test on UR5 30 times to push the seen and unseen cups respectively. The seen and unseen cups vary in shape and color, as shown in Fig. \ref{cups}. And the positions of cups are sampled from the uniform distribution in the representation space. 
The success trajectory is shown in Fig. \ref{pic-traj-real} and the quantitative results are shown in \ref{table2}.
\begin{table}[t]\footnotesize
	\caption{Quantitative results in the real world}
	\vspace{-0.4cm}
	\label{table2}
	\begin{center}
		\linespread{1.2}\selectfont
		\begin{tabular}{c|c|c}
			
			\Xhline{1.5pt}
			
			\multirow{2}{*}{Method} & \multicolumn{2}{c}{Success rate(\%)}  \\
			\cline{2-3}
			& seen cups & unseen cups\\
			
			\Xhline{1.1pt}
			DR-GRL & 73.33 & 70.00 \\
			\hline
			
			\Xhline{1.5pt}
		\end{tabular}
	\end{center}
\end{table}
As is demonstrated, our DR-GRL achieve success rate of 73.33\% to push seen cups and 70.00\% to push unseen cups.  
Since disentangling the appearance and position representation, DR-GRL can focus on the position attributes of objects and generalize to new objects. 

\section{CONCLUSIONS}

We propose a skill learning framework DR-GRL, which learns an object-oriented disentangled representation in a weakly supervised manner to improve the efficiency and generalization of goal-conditioned RL. We design a spatial transform autoencoder (STAE) to learn an interpretable and controllable representation in which different parts correspond to the object attributes (shape, color, position). The representation maintains consistency with the physical positions, which facilitates efficient reward calculation. By simply recoding the learned position representation, STAE can render unseen objects at any position of the image. As a result, the agent can imagine diverse goals to practice itself.
We empirically demonstrate the effectiveness of our DR-GRL. In addition, the framework can conveniently transfer the control policy from simulation to the real world. 

For the task of pushing cups, we can accomplish the task only by focusing on the cup attributes. In the future, we may consider more complex tasks that require attention to the body movement and objects simultaneously, which requires the method to disentangle the manipulator poses into the representation. In addition, how to disentangle multiple objects in more complex scenes is an unsolved issue. Perhaps investigating unsupervised detection of objects in turn or exploiting sequential information would be beneficial.




\bibliographystyle{IEEEtran}     
\bibliography{reference}

\begin{thebibliography}{10}
\providecommand{\url}[1]{#1}
\csname url@samestyle\endcsname
\providecommand{\newblock}{\relax}
\providecommand{\bibinfo}[2]{#2}
\providecommand{\BIBentrySTDinterwordspacing}{\spaceskip=0pt\relax}
\providecommand{\BIBentryALTinterwordstretchfactor}{4}
\providecommand{\BIBentryALTinterwordspacing}{\spaceskip=\fontdimen2\font plus
\BIBentryALTinterwordstretchfactor\fontdimen3\font minus
  \fontdimen4\font\relax}
\providecommand{\BIBforeignlanguage}[2]{{%
\expandafter\ifx\csname l@#1\endcsname\relax
\typeout{** WARNING: IEEEtran.bst: No hyphenation pattern has been}%
\typeout{** loaded for the language `#1'. Using the pattern for}%
\typeout{** the default language instead.}%
\else
\language=\csname l@#1\endcsname
\fi
#2}}
\providecommand{\BIBdecl}{\relax}
\BIBdecl

\bibitem{Mnih2015HumanlevelCT}
V.~Mnih, K.~Kavukcuoglu, D.~Silver, A.~A. Rusu, J.~Veness, M.~G. Bellemare,
  A.~Graves, M.~A. Riedmiller, A.~Fidjeland, G.~Ostrovski, S.~Petersen,
  C.~Beattie, A.~Sadik, I.~Antonoglou, H.~King, D.~Kumaran, D.~Wierstra,
  S.~Legg, and D.~Hassabis, ``Human-level control through deep reinforcement
  learning,'' \emph{Nature}, vol. 518, pp. 529--533, 2015.

\bibitem{Silver2016MasteringTG}
D.~Silver, A.~Huang, C.~J. Maddison, A.~Guez, L.~Sifre, G.~V.~D. Driessche,
  J.~Schrittwieser, I.~Antonoglou, V.~Panneershelvam, M.~Lanctot, S.~Dieleman,
  D.~Grewe, J.~Nham, N.~Kalchbrenner, I.~Sutskever, T.~Lillicrap, M.~Leach,
  K.~Kavukcuoglu, T.~Graepel, and D.~Hassabis, ``Mastering the game of go with
  deep neural networks and tree search,'' \emph{Nature}, vol. 529, pp.
  484--489, 2016.

\bibitem{Silver2017MasteringTG}
D.~Silver, J.~Schrittwieser, K.~Simonyan, I.~Antonoglou, A.~Huang, A.~Guez,
  T.~Hubert, L.~baker, M.~Lai, A.~Bolton, Y.~Chen, T.~Lillicrap, F.~Hui,
  L.~Sifre, G.~V.~D. Driessche, T.~Graepel, and D.~Hassabis, ``Mastering the
  game of go without human knowledge,'' \emph{Nature}, vol. 550, pp. 354--359,
  2017.

\bibitem{Schrittwieser2020MasteringAG}
J.~Schrittwieser, I.~Antonoglou, T.~Hubert, K.~Simonyan, L.~Sifre, S.~Schmitt,
  A.~Guez, E.~Lockhart, D.~Hassabis, T.~Graepel, T.~Lillicrap, and D.~Silver,
  ``Mastering atari, go, chess and shogi by planning with a learned model,''
  \emph{Nature}, vol. 588 7839, pp. 604--609, 2020.

\bibitem{Su2020ReinforcementLB}
H.~Su, Y.~Hu, Z.~Li, A.~Knoll, G.~Ferrigno, and E.~D. Momi, ``Reinforcement
  learning based manipulation skill transferring for robot-assisted minimally
  invasive surgery,'' \emph{2020 IEEE International Conference on Robotics and
  Automation (ICRA)}, pp. 2203--2208, 2020.

\bibitem{Yang2020RigidSoftIL}
L.~Yang, F.~Wan, H.~Wang, X.~Liu, Y.~Liu, J.~Pan, and C.~Song, ``Rigid-soft
  interactive learning for robust grasping,'' \emph{IEEE Robotics and
  Automation Letters}, vol.~5, pp. 1720--1727, 2020.

\bibitem{Pedersen2020GraspingUO}
O.-M. Pedersen, E.~Misimi, and F.~Chaumette, ``Grasping unknown objects by
  coupling deep reinforcement learning, generative adversarial networks, and
  visual servoing,'' \emph{2020 IEEE International Conference on Robotics and
  Automation (ICRA)}, pp. 5655--5662, 2020.

\bibitem{Nair2018VisualRL}
A.~Nair, V.~H. Pong, M.~Dalal, S.~Bahl, S.~Lin, and S.~Levine, ``Visual
  reinforcement learning with imagined goals,'' in \emph{NeurIPS}, 2018.

\bibitem{Srinivas2020CURLCU}
M.~Laskin, A.~Srinivas, and P.~Abbeel, ``Curl: Contrastive unsupervised
  representations for reinforcement learning,'' in \emph{International
  Conference on Machine Learning}.\hskip 1em plus 0.5em minus 0.4em\relax PMLR,
  2020, pp. 5639--5650.

\bibitem{Feng2018DualSD}
Z.~Feng, X.~Wang, C.~Ke, A.~Zeng, D.~Tao, and M.~Song, ``Dual swap
  disentangling,'' in \emph{NeurIPS}, 2018.

\bibitem{ge2021zeroshot}
\BIBentryALTinterwordspacing
Y.~Ge, S.~Abu-El-Haija, G.~Xin, and L.~Itti, ``Zero-shot synthesis with
  group-supervised learning,'' in \emph{International Conference on Learning
  Representations}, 2021. [Online]. Available:
  \url{https://openreview.net/forum?id=8wqCDnBmnrT}
\BIBentrySTDinterwordspacing

\bibitem{Niemeyer2020GIRAFFE}
M.~Niemeyer and A.~Geiger, ``Giraffe: Representing scenes as compositional
  generative neural feature fields,'' in \emph{Proc. IEEE Conf. on Computer
  Vision and Pattern Recognition (CVPR)}, 2021.

\bibitem{Schaul2015UniversalVF}
T.~Schaul, D.~Horgan, K.~Gregor, and D.~Silver, ``Universal value function
  approximators,'' in \emph{ICML}, 2015.

\bibitem{Andrychowicz2017HindsightER}
M.~Andrychowicz, F.~Wolski, A.~Ray, J.~Schneider, R.~Fong, P.~Welinder,
  B.~McGrew, J.~Tobin, P.~Abbeel, and W.~Zaremba, ``Hindsight experience
  replay,'' in \emph{Proceedings of the 31st International Conference on Neural
  Information Processing Systems}, 2017, pp. 5055--5065.

\bibitem{Kingma2014AutoEncodingVB}
D.~P. Kingma and M.~Welling, ``Auto-encoding variational bayes,'' \emph{stat},
  vol. 1050, p.~1, 2014.

\bibitem{Nair2019ContextualIG}
A.~Nair, S.~Bahl, A.~Khazatsky, V.~H. Pong, G.~Berseth, and S.~Levine,
  ``Contextual imagined goals for self-supervised robotic learning,'' in
  \emph{CoRL}, 2019.

\bibitem{Oord2017NeuralDR}
A.~van~den Oord, O.~Vinyals, and K.~Kavukcuoglu, ``Neural discrete
  representation learning,'' in \emph{Proceedings of the 31st International
  Conference on Neural Information Processing Systems}, 2017, pp. 6309--6318.

\bibitem{Khazatsky2021WhatCI}
A.~Khazatsky, A.~Nair, D.~Jing, and S.~Levine, ``What can i do here? learning
  new skills by imagining visual affordances,'' \emph{2021 IEEE International
  Conference on Robotics and Automation (ICRA)}, pp. 14\,291--14\,297, 2021.

\bibitem{Wang2020ROLLVS}
Y.~Wang, G.~N. Narasimhan, X.~Lin, B.~Okorn, and D.~Held, ``Roll: Visual
  self-supervised reinforcement learning with object reasoning,'' in
  \emph{CoRL}, 2020.

\bibitem{Park2020LearningTA}
J.~H. Park, J.~Kim, Y.~Jang, I.~Jang, and H.~Kim, ``Learning transformable and
  plannable se(3) features for scene imitation of a mobile service robot,''
  \emph{IEEE Robotics and Automation Letters}, vol.~5, pp. 1664--1671, 2020.

\bibitem{Pathak2017LearningFB}
D.~Pathak, R.~B. Girshick, P.~Doll{\'a}r, T.~Darrell, and B.~Hariharan,
  ``Learning features by watching objects move,'' \emph{2017 IEEE Conference on
  Computer Vision and Pattern Recognition (CVPR)}, pp. 6024--6033, 2017.

\bibitem{kulkarni2019unsupervised}
T.~D. Kulkarni, A.~Gupta, C.~Ionescu, S.~Borgeaud, M.~Reynolds, A.~Zisserman,
  and V.~Mnih, ``Unsupervised learning of object keypoints for perception and
  control,'' \emph{Advances in Neural Information Processing Systems}, vol.~32,
  pp. 10\,724--10\,734, 2019.

\bibitem{zhang2020learning}
A.~Zhang, R.~T. McAllister, R.~Calandra, Y.~Gal, and S.~Levine, ``Learning
  invariant representations for reinforcement learning without
  reconstruction,'' in \emph{International Conference on Learning
  Representations}, 2021.

\bibitem{chen2018isolating}
R.~T. Chen, X.~Li, R.~Grosse, and D.~Duvenaud, ``Isolating sources of
  disentanglement in vaes,'' in \emph{Proceedings of the 32nd International
  Conference on Neural Information Processing Systems}, 2018, pp. 2615--2625.

\bibitem{zhang2021sill}
H.~Zhang, Z.~Cao, Z.~Yan, and C.~Zhang, ``Sill-net: Feature augmentation with
  separated illumination representation,'' \emph{arXiv preprint
  arXiv:2102.03539}, 2021.

\bibitem{niemeyer2021giraffe}
M.~Niemeyer and A.~Geiger, ``Giraffe: Representing scenes as compositional
  generative neural feature fields,'' in \emph{Proceedings of the IEEE/CVF
  Conference on Computer Vision and Pattern Recognition}, 2021, pp.
  11\,453--11\,464.

\bibitem{ge2020zero}
Y.~Ge, S.~Abu-El-Haija, G.~Xin, and L.~Itti, ``Zero-shot synthesis with
  group-supervised learning,'' in \emph{International Conference on Learning
  Representations}, 2020.

\bibitem{zhang2018mixup}
H.~Zhang, M.~Cisse, Y.~N. Dauphin, and D.~Lopez-Paz, ``mixup: Beyond empirical
  risk minimization,'' in \emph{International Conference on Learning
  Representations}, 2018.

\bibitem{Zivkovic2006EfficientAD}
Z.~Zivkovic and F.~van~der Heijden, ``Efficient adaptive density estimation per
  image pixel for the task of background subtraction,'' \emph{Pattern Recognit.
  Lett.}, vol.~27, pp. 773--780, 2006.

\bibitem{bradski2008learning}
G.~Bradski and A.~Kaehler, \emph{Learning OpenCV: Computer vision with the
  OpenCV library}.\hskip 1em plus 0.5em minus 0.4em\relax " O'Reilly Media,
  Inc.", 2008.

\bibitem{10.5555/2969442.2969465}
M.~Jaderberg, K.~Simonyan, A.~Zisserman, and K.~Kavukcuoglu, ``Spatial
  transformer networks,'' in \emph{Proceedings of the 28th International
  Conference on Neural Information Processing Systems - Volume 2}, ser.
  NIPS'15.\hskip 1em plus 0.5em minus 0.4em\relax Cambridge, MA, USA: MIT
  Press, 2015, p. 2017–2025.

\bibitem{haarnoja2018soft}
T.~Haarnoja, A.~Zhou, P.~Abbeel, and S.~Levine, ``Soft actor-critic: Off-policy
  maximum entropy deep reinforcement learning with a stochastic actor,'' in
  \emph{International conference on machine learning}.\hskip 1em plus 0.5em
  minus 0.4em\relax PMLR, 2018, pp. 1861--1870.

\bibitem{10.1007/978-3-319-24574-4_28}
O.~Ronneberger, P.~Fischer, and T.~Brox, ``U-net: Convolutional networks for
  biomedical image segmentation,'' in \emph{Medical Image Computing and
  Computer-Assisted Intervention -- MICCAI 2015}.\hskip 1em plus 0.5em minus
  0.4em\relax Cham: Springer International Publishing, 2015, pp. 234--241.

\bibitem{2018CBAM}
S.~Woo, J.~Park, J.~Y. Lee, and I.~S. Kweon, ``Cbam: Convolutional block
  attention module,'' in \emph{European Conference on Computer Vision}, 2018.

\bibitem{Maaten2008VisualizingDU}
L.~V.~D. Maaten and G.~E. Hinton, ``Visualizing data using t-sne,''
  \emph{Journal of Machine Learning Research}, vol.~9, pp. 2579--2605, 2008.

\bibitem{Zhang2018TheUE}
R.~Zhang, P.~Isola, A.~A. Efros, E.~Shechtman, and O.~Wang, ``The unreasonable
  effectiveness of deep features as a perceptual metric,'' \emph{2018 IEEE/CVF
  Conference on Computer Vision and Pattern Recognition}, pp. 586--595, 2018.

\end{thebibliography}


%



\end{document}